\documentclass[sigconf]{acmart-me}

\usepackage{booktabs} 
\usepackage{url}
\usepackage{color}
\usepackage{enumitem}
\usepackage{subcaption}
\usepackage{balance}

\setcopyright{rightsretained}

\acmDOI{}

\acmISBN{}

\begin{document}
\title{Multi-Modal Machine Learning for Flood Detection in News, Social Media and Satellite Sequences}

\author{Kashif Ahmad\textsuperscript{1}, Konstantin Pogorelov \textsuperscript{2}, Mohib Ullah\textsuperscript{3},\\ Michael Riegler\textsuperscript{4}, Nicola Conci\textsuperscript{5}, Johannes Langguth \textsuperscript{2}, Ala Al-Fuqaha \textsuperscript{1}}
\affiliation{\textsuperscript{1}Hamad Bin Khalifa University, Doha, Qatar\\ \textsuperscript{2} Simula Research Laboratory, Norway \\ \textsuperscript{3}Norwegian University of Science and Technology, Norway \\ \textsuperscript{4} Simula Metropolitan Center for Digitalisation and Kristiania University College, Norway \\ \textsuperscript{5} University of Trento, Italy}
\email{kahmad@hbku.edu.qa, konstantin@simula.no, mohib.ullah@ntnu.no}
\email{michael@simula.no, nicola.conci@unitn.it, langguth@simula.no, aalfuqaha@hbku.edu.qa}

%
%
%
%
%

\renewcommand{\shortauthors}{K. Ahmad et al.}
\renewcommand{\shorttitle}{}

\begin{abstract}
In this paper we present our methods for the MediaEval 2019 Multimedia Satellite Task, which is aiming to extract complementary information associated with adverse events from Social Media and satellites. For the first challenge, we propose a framework jointly utilizing colour, object and scene-level information to predict whether the topic of an article containing an image is a flood event or not. Visual features are combined using early and late fusion techniques achieving an average F1-score of $82.63$, $82.40$, $81.40$ and  $76.77$.
For the multi-modal flood level estimation, we rely on both visual and textual information achieving an average F1-score of $58.48$ and $46.03$, respectively. Finally, for the flooding detection in time-based satellite image sequences we used a combination of classical computer-vision and machine learning approaches achieving an average F1-score of $58.82\%$. 
\end{abstract}

%
%
%
%
%


\maketitle

\section{Introduction}
\label{sec:intro}
When natural disasters occur, an instant access to relevant information may be crucial to mitigate loss in terms of property and human lives, and may result in a speedy recovery \cite{said2019natural}. In this regards, social media and remotely sensed information have been proved very effective \cite{said2019natural,ahmad2019automatic,bischke2016contextual}. 
Similar to the 2017 \cite{bischke2017multimedia} and 2018 \cite{bischke2018multimediasatellite} versions of the task, the MediaEVal 2019 Multimedia Satellite task \cite{bischke2019mmsat} aims to combine the information from the two complementary sources, namely social media and satellites. 

This paper provides detailed description of the methods proposed by team UTAOS for the MediaEval 2019 Multimedia Satellite challenge. The challenge consists of three parts, namely (i) News Image Topic Disambiguation (NITD), (ii) Multimodal Flood Level Estimation (MFLE) and (iii) City-centered Satellite Sequences (CSS). 

The first two tasks(i and ii) are based on social media data aiming to (a) predict whether the topic of the article containing the image was a water-related natural-disaster event or not, and (b) to build a binary classifier that predicts whether or not the image contains at least one person standing in water above the knee. 

In the final task(iii), the participants are provided with a set of sequences of satellite images depicting a certain city over a certain length of time, and the they need to propose and develop a framework able to determine whether or not there was a flooding event ongoing in that city at that time.

\section{Proposed Approach}
\subsection{Methodology for NITD task}
Considering the diversity of the content covered by natural disaster-related images, based on our previous experience \cite{ahmad2018comparative}, we utilize a diversified set of visual features including colour, object and scene-level features. The object and scene-level features are extracted through through three different Convolutional Neural Network (CNN) models, namely AlexNet\cite{krizhevsky2012imagenet}, VggNet \cite{simonyan2014very} and  ResNet \cite{he2016deep}, pre-trained on the ImageNet dataset \cite{deng2009imagenet} and the Places dataset \cite{zhou2014learning}. The models pre-trained on Imagenet correspond to object level information while the ones pre-trained on the Places dataset extract scene level information. For feature extraction from all models, we use the Caffe toolbox\footnote{http://caffe.berkeleyvision.org/}. 
For colour and texture features we rely on the LIRE  open source library \cite{lux2016lire} which we used to extract joint composite descriptor (JCD) features from the images. 

In order to combine the features, we use both early and late fusion techniques. For the early fusion, feature vectors are concatenated.
For late fusion two different techniques namely (i) simple averaging and (ii) Particle Swarm Optimization (PSO) based technique is used for late fusion. For classification purposes, we rely on Support Vector Machines (SVMs) in all of the submitted fusion runs. Moreover, to deal with class imbalance problem, we use ensemble different re-sampled data sets technique where five different models are trained using all the samples of the rare class (i.e., class 0) and n-differing samples of the abundant class (i.e., class 1).

\subsection{Methodology for MFLE task}
For the MFLE task, we proposed two different solutions exploiting both: visual and textual information. For visual features based flood estimation, we proposed a two step framework where as a first step a binary image classifier based on deep visual features is used to differentiate between flood and non-flooded images. In the second step, an open source library, namely OpenPose\footnote{https://www.learnopencv.com/tag/openpose/}, has been used to draw and extract body points on the people in the flood related images. Subsequently, the generated coordinates are analyzed to identify the images having at least one person standing in water and the water level is above the knee height by targeting the knee joints. On the other hand, for text analysis we employed two methods, namely (i) Bag-of-words Model (BoW) and (ii) LSTM network. Before applying the methods, the data was pre-processed for tokenization and removing of punctuation.

\subsection{Methodology for CSS task}
For CSS task, first, we tried to employ a recurrent convolutional neural network architecture designed for change detection in multi-spectral satellite imagery data (ReCNN)~\cite{mou2018learning}. This network was initially designed to solve the task very similar to CSS task goals, and the results depicted by ReCNN's authors are promising. However, despite high expectations, ReCNN was not able to achieve sufficient and better-than-random performance of detection changes caused by flooding. Our assumption is that was caused by the "real" nature of the dataset provided in CSS task. Images were taken in different seasons, often partially or fully covered by clouds and sometimes have noticeable pixel offset between each other. After a series of unsuccessful experiments, we decided to use a classical image processing and analysis approach with multi-stage image processing using simple operations. First, we mask-out from the further analysis all cloud-covered image areas by applying simple threshold function. Reference threshold value is computed per-image by averaging the values of the pixels located in monotonically white-colored areas. The same masking is performed for dark underexposured and areas with missing imaging data. Next, we scaled images down to uniform size of $128x128$ pixels to reduce noise and soften image-shifting influence. Then, scaled images are converted into hue-saturation-value (HSV) color space, and further analysis is performed on HSV bands. Using the same thresholding methodology, we mask-out pixels with too-low and too-high S and V channel values. The resulting masks are filtered with median filter and processed by dilation filter. Resulting images are compared in sequential pairs within non-masked-out regions using grey level co-occurrence matrix texture feature. Final flooding presence detection is made by using random tree classifier.

\section{Results and Analysis}

\subsection{Runs Description in NITD Task}
For NITD, we submitted total four runs. In run 1, we used the PSO based weight optimization method for assigning weights to each model on merit basis. For run 2, the deep models are treated equally by assigning equal weights to all models. In our run 3, we added colour based features to our pool of features descriptors in a late fusion method where the scores of all models are simply added to obtain the final prediction. Our run 4 is based early fusion where the deep features are simply concatenated for training SVMs. Table \ref{NITD_results} provides the experimental results our proposed solutions for NITD task. Overall better results are obtained with PSO based late fusion which shows the advantage of merit based late fusion of the models. On the other hand, least F1-score is obtained with early fusion. Moreover, the colour based features did not contribute positively in the performance of the framework. This might be due to the fact that the JCD feature is very compressed and does not contain much information that the fusion algorithm could exploit.

\begin{table}[!htb]
    \caption{Evaluation of our proposed approach for (a) NITD and (b) MLFE task in terms of F1-scores.}
    \vspace{-10px}
    \begin{subtable}{.5\linewidth}
      \centering
        \caption{NITD}
\label{NITD_results}
\begin{tabular}{|c|c|c|c|}
\hline
\textbf{Run} & \textbf{F1-score} \\ \hline
Run 1 & 82.63  \\ \hline
Run 2 & 82.40  \\ \hline
Run 3 & 81.40  \\ \hline
Run 4 & 76.77   \\ \hline
\end{tabular}
    \end{subtable}%
    \begin{subtable}{.5\linewidth}
      \centering
        \caption{MFLE}
\label{MFLE_results}
\begin{tabular}{|c|c|}
\hline
\textbf{Run} & \textbf{F1-score} \\ \hline
Run 1  &   58.48  \\ \hline
Run 2  &  46.03   \\ \hline
Run 4  &  44.91   \\ \hline
\end{tabular}
    \end{subtable} 
    \vspace{-10px}
\end{table}

\subsection{Runs Description in MLFE Task}
For MLFE task, we submitted two mandatory and one optional run. The first run is based on visual information where a two phase approach has been proposed for flood level estimation starting with deep features based classification of flooded and non-flooded images, followed by human body points detection via Openpose library in the flood-related images. Our second and third runs are based on textual information where Bag-of-words (BoW) and LSTM based techniques are used for the article classification, respectively.  Table \ref{MFLE_results} shows the experimental results of our solutions for MLFE task. Overall, better results are obtained with visual information. Moreover, BoW features produce slightly better results over LSTM based approach. 

\subsection{Runs Description in CSS Task}

For CSS task we submitted the mandatory run only. Evaluation performed by the task organizers showed F1-score of $58.82\%$ for flooding detection performance on the provided test set. The relatively high performance for our simple detection approach can be explained by the used aggressive image masking technique which allow us to perform comparison of only clearly visible areas. However, our own evaluation shows that our approach is not able to distinguish correctly between image changes caused by flooding and seasonal vegetation grow.

\section{Conclusions and Future Work}
This year, the social multimedia satellite  task introduced a new and important challenges including image based news topic disambiguation (NITD), multi-modal flood level estimation in social media content (MFLE) and predicting a flood event in a set of sequences of satellite images of a certain city over a certain length of time (CSS). For the NITD task, we mainly relied on ensembles of classifiers trained on deep features extracted through several pre-trained deep models as well as global features (GF). During the experiments, we observed that the object and scene-level features complement each others when jointly utilized in a proper way. Moreover, deep features are proved more effective compared to GF. For MLFE task, we used both textual and visual information where better results were obtained with visual information. However, textual and visual information can complement each other. In the future, we aim to analyze the task with more advanced early and late fusion techniques to better utilize the multi-modal information. Furthermore, we plan to use complex GF. For CSS task, we used the combination of computer-vision and machine learning approaches. For future results improvement, we will continue investigating recurrent CNN and GAN-based approaches in combination with classical image processing algorithms.

\balance

\bibliographystyle{ACM-Reference-Format}
\def\bibfont{\small} 
\bibliography{sample-me19} 

\end{document}